\pdfoutput=1
\documentclass[11pt,table,xcdraw]{article}

\newcommand{\single}{\textsc{Single Turnback }}
\newcommand{\return}{\textsc{Return Turnback }}
\newcommand{\dualval}{\textsc{Dual-value Turnback }}
\newcommand{\dualslot}{\textsc{Dual-slot Turnback }}

\usepackage{acl}
\usepackage{kotex}
\usepackage{tabularx}
\usepackage{booktabs}
\usepackage{boldline}
\usepackage{subcaption}
\usepackage{multirow}
\usepackage{float}
\usepackage{xcolor}
\usepackage{times}
\usepackage{latexsym}

\usepackage{graphicx}

\usepackage[T1]{fontenc}

\usepackage[utf8]{inputenc}

\usepackage{microtype}

\title{Oh My Mistake!: Toward Realistic \\ Dialogue State Tracking including Turnback Utterances}

\author{Takyoung Kim$^{\dagger}$, Yukyung Lee$^{\dagger}$, Hoonsang Yoon$^{\dagger}$, \\ \textbf{Pilsung Kang}$^{\dagger}$, \textbf{Junseong Bang}$^{\ddagger}$, \textbf{Misuk Kim}$^{\mathsection}$ \\
        Korea University, Seoul 02841, Republic of Korea$^\dagger$ \\ 
        Electronics and Telecommunications Research Institute, Daejeon 34129, Republic of Korea$^{\ddagger}$ \\
        Sejong University, Seoul 05006, Republic of Korea$^\mathsection$ \\
        \texttt{\small \{takyoung\_kim, yukyung\_lee, hoonsang\_yoon, pilsung\_kang\}@korea.ac.kr}\\ 
        \texttt{\small misuk.kim@sejong.ac.kr}}

\begin{document}
\maketitle
\begin{abstract}
The primary purpose of dialogue state tracking (DST), a critical component of an end-to-end conversational system, is to build a model that responds well to real-world situations. Although we often change our minds from time to time during ordinary conversations, current benchmark datasets do not adequately reflect such occurrences and instead consist of over-simplified conversations, in which no one changes their mind during a conversation. As the main question inspiring the present study, ``Are current benchmark datasets sufficiently diverse to handle casual conversations in which one changes their mind after a certain topic is over?'' We found that the answer is ``No'' because DST models cannot refer to previous user preferences when template-based turnback utterances are injected into the dataset. Even in the the simplest mind-changing (turnback) scenario, the performance of DST models significantly degenerated. However, we found that this performance degeneration can be recovered when the turnback scenarios are explicitly designed in the training set, implying that the problem is not with the DST models but rather with the construction of the benchmark dataset.
\end{abstract}

%%%%%%%%%%%%%%%%%%%%%%%%%%%%%%%%%%%%%%%%%
%  1. Introduction                      %
%%%%%%%%%%%%%%%%%%%%%%%%%%%%%%%%%%%%%%%%%
\section{Introduction}

\begin{figure*}[h!]
    \centering
    \begin{subfigure}[b]{0.45\textwidth}
        \centering
        \includegraphics[width=\textwidth]{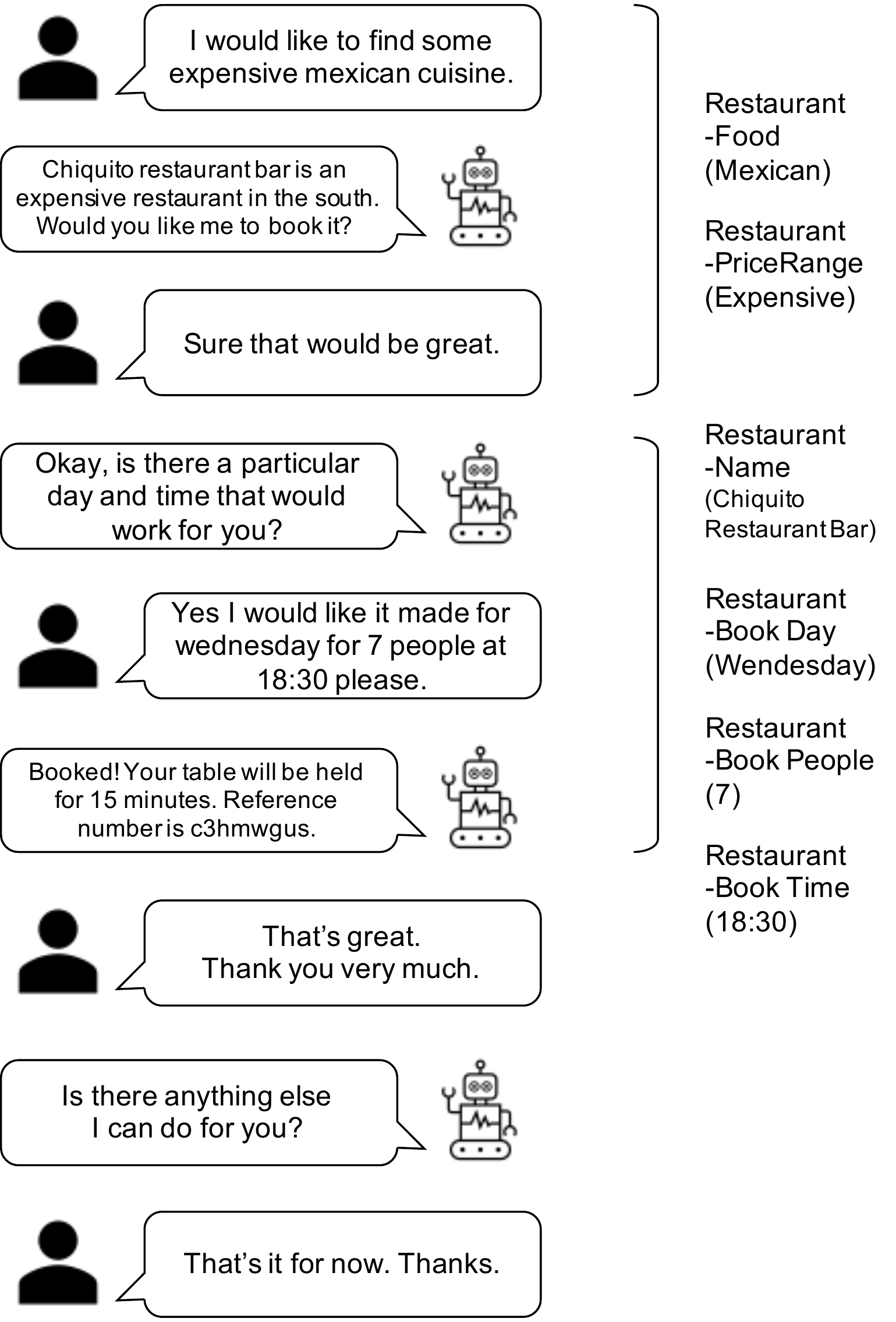}
        \caption{Benchmark}
        \label{fig:1a}
    \end{subfigure}
    % \hspace{0.5cm}
    \begin{subfigure}[b]{0.45\textwidth}  
        \centering 
        \includegraphics[width=\textwidth]{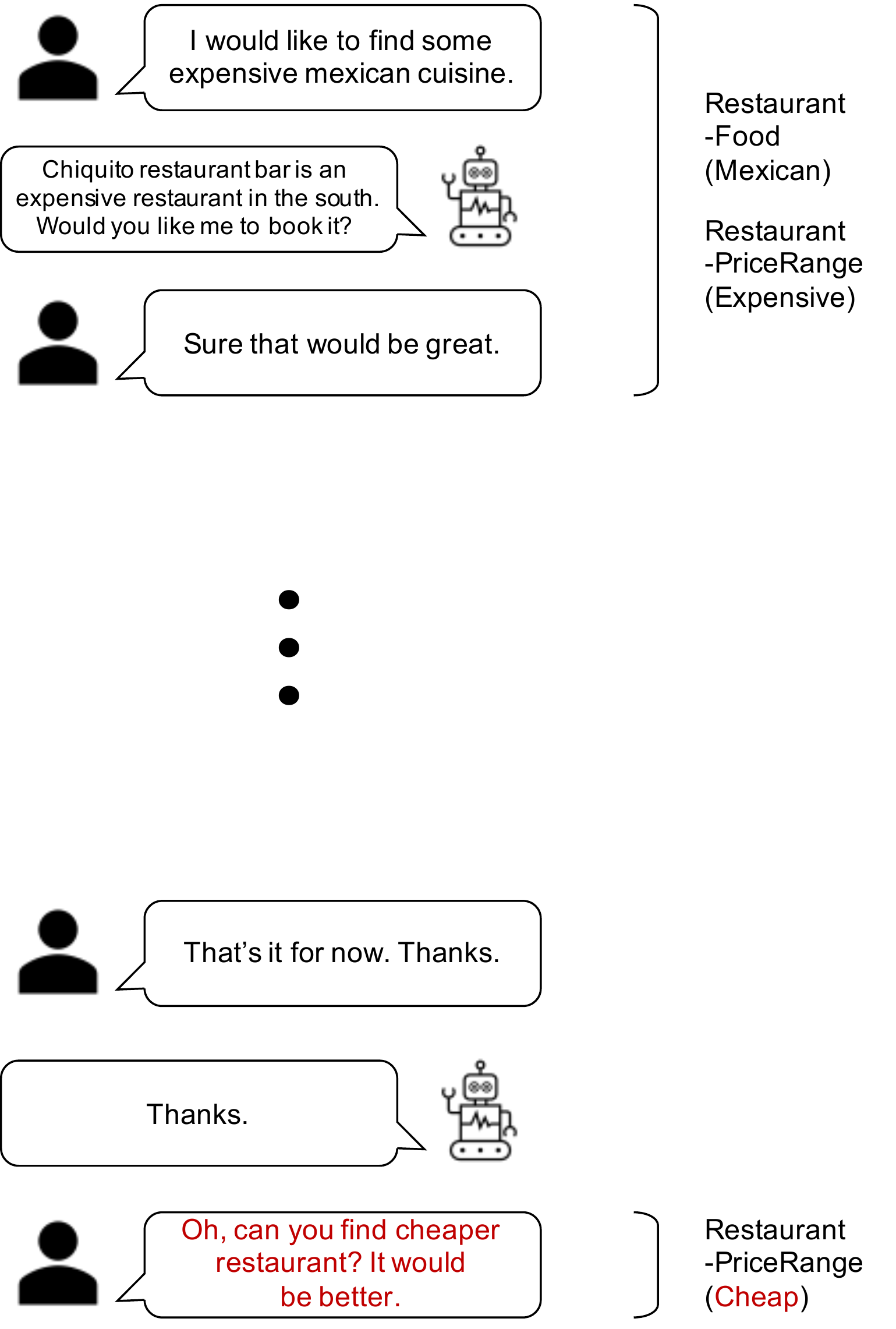}
        \caption{In reality}
        \label{fig:1b}
    \end{subfigure}
    \caption
    {Dialogue flow example of MultiWOZ 2.1 (MUL1514.json).} 
    \label{fig:1}
\end{figure*}

The dialogue state tracking (DST) module is a part of a task-oriented dialogue system, the main role of which is to extract essential information of user preferences from various conversational situations. Based on the given information from the previous module, the DST module finds appropriate slot-value pairs to understand the current conversational situations, and these pairs are then delivered to the next module to continue the conversation. Hence, building an accurate DST model is a key success factor of the overall task-oriented dialogue system not only because it can convince users that the system perfectly understands what they are talking about, but also because appropriate responses can be generated based on the result of the DST model. As in other natural language processing (NLP) tasks, two main components are mandatory to build a good DST model: (1) well-structured machine learning models and (2) sufficiently large datasets that contain various real-world conversational situations with fewer biases for training the model. Since the introduction of Transformer and BERT \cite{vaswani2017attention, devlin2018ert}, various breakthrough model structures have been designed for DST, such as SUMBT and SOM-DST \cite{lee2019sumbt, kim2019efficient}, and have shown an excellent performance. With respect to DST-specific datasets, by contrast, some benchmark datasets, such as WOZ \cite{wen2017network} and MultiWOZ \cite{budzianowski2018large}, have been introduced; however, their sizes and coverage are not yet satisfactory owing to the relatively high labeling cost. For example, the MultiWOZ only consists of approximately 10,000 dialogues from some different domains, which is significantly smaller than other NLP 
datasets such as SQuAD or IMDB \cite{rajpurkar-etal-2016-squad, maas-etal-2011-learning}.

Whereas the MultiWOZ has been used as a standard benchmark dataset for DST, there has been an increasing number of recent studies reporting the concerns regarding the inherent limitations of this dataset. First, newer versions of MultiWOZ have been proposed to address certain issues such as annotation errors, typos, standardization, annotation consistency, and other factors \cite{eric2019multiwoz, zang2020multiwoz, han2020multiwoz, ye2021multiwoz}. In addition, \citet{qian2021annotation} pointed out an entity bias issue, i.e., only a small number of values in the ontology account for the majority of labels. For example, a large number of \textit{`train-destination'} slots take the value \textit{`cambridge'} in the MultiWOZ \cite{qian2021annotation}. In addition, with CoCo \cite{li2020coco}, an overestimation of the held-out accuracy was pointed out by showing that the training and evaluation sets of the MultiWOZ have a similar distribution, and controllable counterfactual goals were proposed that do not change the original dialogue flow but generate a new dialogue with different responses.

Although previous studies have raised inherent problems in the MultiWOZ, most have tended to focus on correcting the annotation inconsistency or entity biases, which enforces the dialogue in the dataset to be more idealistic. However, in real-world conversations, the dialogue flow between two speakers is not always as fluent as those in the MultiWOZ, e.g., one can occasionally change one's mind during a conversation. For example, Figure \ref{fig:1a} shows a sample dialogue in the MultiWOZ. No slot that appears once appears again in the subsequent dialogue turns. As the main hypothesis motivating this study, real conversations do not always continue as shown in Figure \ref{fig:1a}, but often continue as shown in Figure \ref{fig:1b}. Individuals change their mind during a conversation, and thus some slot-value pairs (same slot but different values) repeatedly appear in an entire dialogue. This hypothesis has led us to raise the main question of this paper: \textit{``Can the current benchmark dataset handle a situation in which users change their mind after a certain amount of turn?''} Our assumption is that the turnback situation of a user will hamper the robust evaluation of DST models because such models do not have a chance to learn the situation in which the values of specific slots are changed during the conversation. To experimentally verify our assumption, we investigate how DST models handle additional turnback dialogues by injecting template-based utterances under different scenarios on the MultiWOZ.

It is common for users to change their decisions in various ways in the real world, and thus we define four turnback situations as follows: 
\begin{itemize}
    \item {\textbf{\single}}: This is the simplest form in which the user changes the decision of a single slot only once.
    \item {\textbf{\return}}: This is the reverse of a decision twice but returning to the original value of a single slot.
    \item {\textbf{\dualval}}: The decision for a single slot is changed twice and thus the corresponding values are also changed twice.
    \item {\textbf{\dualslot}}: The decision for two slots are sequentially changed. The corresponding values are changed only once.
\end{itemize} 
The remaining states are more complicated variants of the simplest versions by modifying the number of repetitions or slots. There are some ways to generate turnback utterances such as manually annotating dialogues or generating with the help of language models \citep{2020t5}. In this study, we injected turnback utterances at the end of the existing dialogue using pre-defined templates for two reasons. First, locating turnback utterances at the end of the dialogue is a better way to verify the ability handling long-range contexts for the model. Second, template-based-generated utterances explicitly mention the information of \texttt{domain}, \texttt{slot}, and \texttt{value} in a raw text, which can play a role as the minimal form of turnback scenarios. We found even these simple and explicit forms of turnback utterances are sufficient to disclose the problem.

In this paper, we evaluate the performance of turnback situations with TRADE, SUMBT, and Transformer-DST \cite{WuTradeDST2019, lee2019sumbt, zeng2020jointly}. The results show that existing models cannot detect changing user preferences when injecting turnback utterances in the test set; the same trends are also shown in all variants of turnback scenarios. We further determined that including turnback utterances appropriately during the training phase can make a model robust because the model performance rebounds. To summarize, the main contributions of this paper can be summarized as follows:

\begin{itemize}
    \item We define the problem that the current benchmark cannot handle, i.e., the change in decision of the user after a certain topic is over, which must be considered when constructing an realistic conversational system.
    \item We quantitatively and qualitatively evaluate three representative DST models to verify the effect of the turnback situation by injecting template-based utterances into the existing dataset. 
    \item We explore the effect of various turnback proportions in both the training and testing datasets: When turnback utterances appear in the test set, models trained with the data including turnback utterances become more robust.
\end{itemize}

\section{Related Work}
\label{related_work}

\subsection{Limitation of Benchmark Dataset}
MultiWOZ \cite{budzianowski2018large} is one of the most popular multi-domain task-oriented dialogue datasets.
Although a new task-oriented dialogue dataset, such as SGD \cite{rastogi2020towards}, has been recently proposed, most previous studies still evaluate the performance based on MultiWOZ \citep{kim-etal-2022-mismatch}. However, it has been revealed that the MultiWOZ has inherent errors and biases, and several studies have been proposed to resolve the reported issues.

\paragraph{Annotation error}
Even the recent versions of MultiWOZ still have incorrect labels and inconsistent annotations \cite{eric2019multiwoz, zang2020multiwoz, han2020multiwoz, ye2021multiwoz}. These noises are the primary reason why it is challenging to accurately evaluate the model performance. Fortunately, the benchmark is continuously updated by progressively correcting any annotation errors found.

\paragraph{Biased slots}
The slots in MultiWOZ are biased. The slots in the training and test sets overlap by more than 90\%, and the co-occurrence between slots in the test set is also unequally distributed. DST models are vulnerable to unseen slots because biased slots do not consider rare but realistic slot combinations. To relieve this assumption, CoCo  \cite{li2020coco} generates counterfactual dialogues to allow the existing dataset to cover realistic conversation scenarios.

\paragraph{Biased entities}
Entities in the MultiWOZ are also significantly biased. The test dataset has most of the entities that appear in the training dataset, and existing models are vulnerable to unseen entities (e.g., \textit{``cambridge''} appearing in 50\% of the destination cities in the \textit{train} domain) \cite{qian2021annotation}. Thus, the new test dataset consisting of unseen entities is proposed, which also results in a decrease in performance \cite{qian2021annotation}.

\begin{figure*}
    \centering
    \begin{subfigure}[b]{0.245\textwidth}
        \centering
        \includegraphics[width=\textwidth]{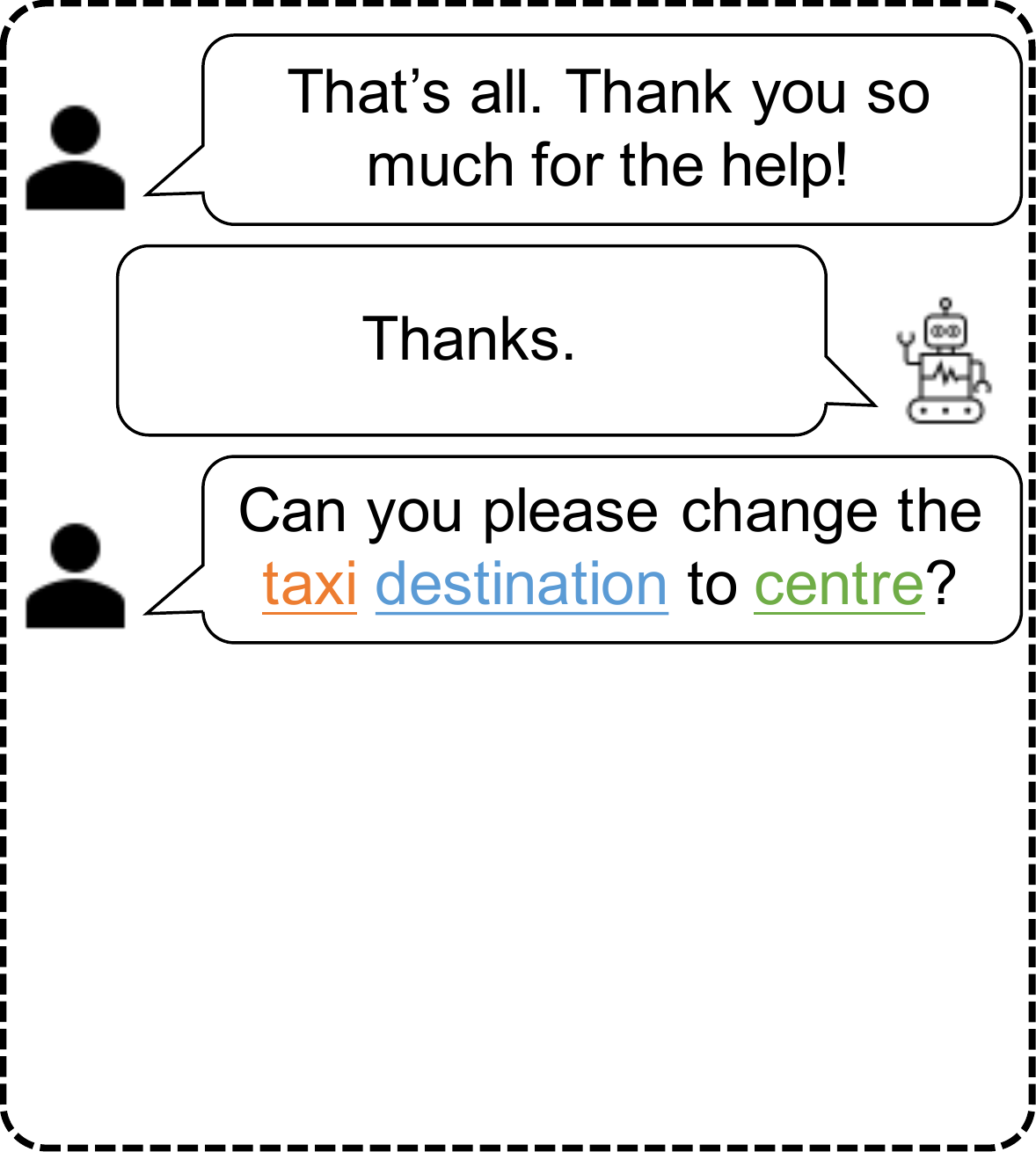}
        \caption{\single}
        \label{fig:3a}
    \end{subfigure}
    % \hspace{0.1cm}
    \begin{subfigure}[b]{0.245\textwidth}  
        \centering 
        \includegraphics[width=\textwidth]{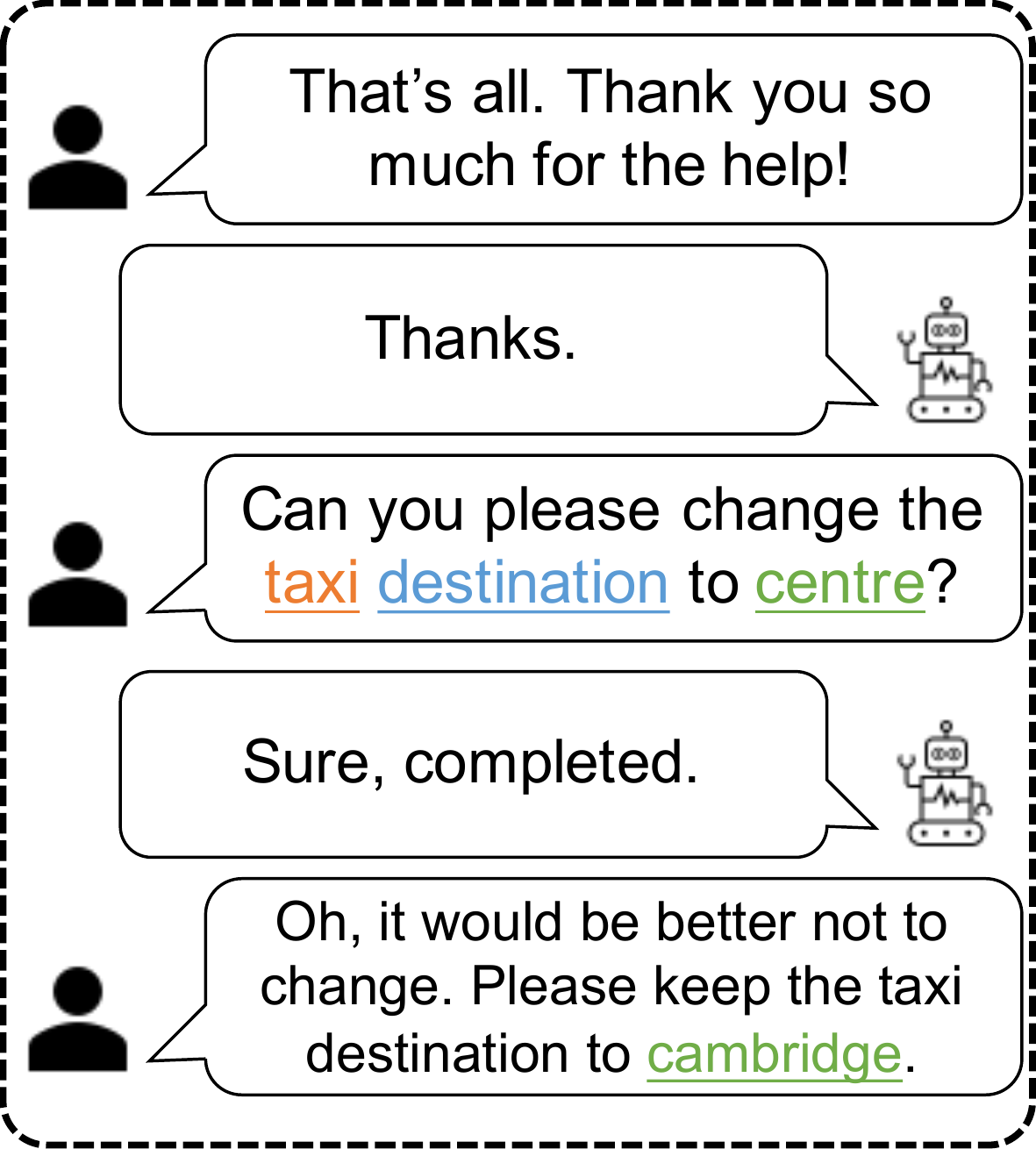}
        \caption{\return}
        \label{fig:3b}
    \end{subfigure}
    % \hspace{0.1cm}
    % \vskip\baselineskip
    \begin{subfigure}[b]{0.245\textwidth}   
        \centering 
        \includegraphics[width=\textwidth]{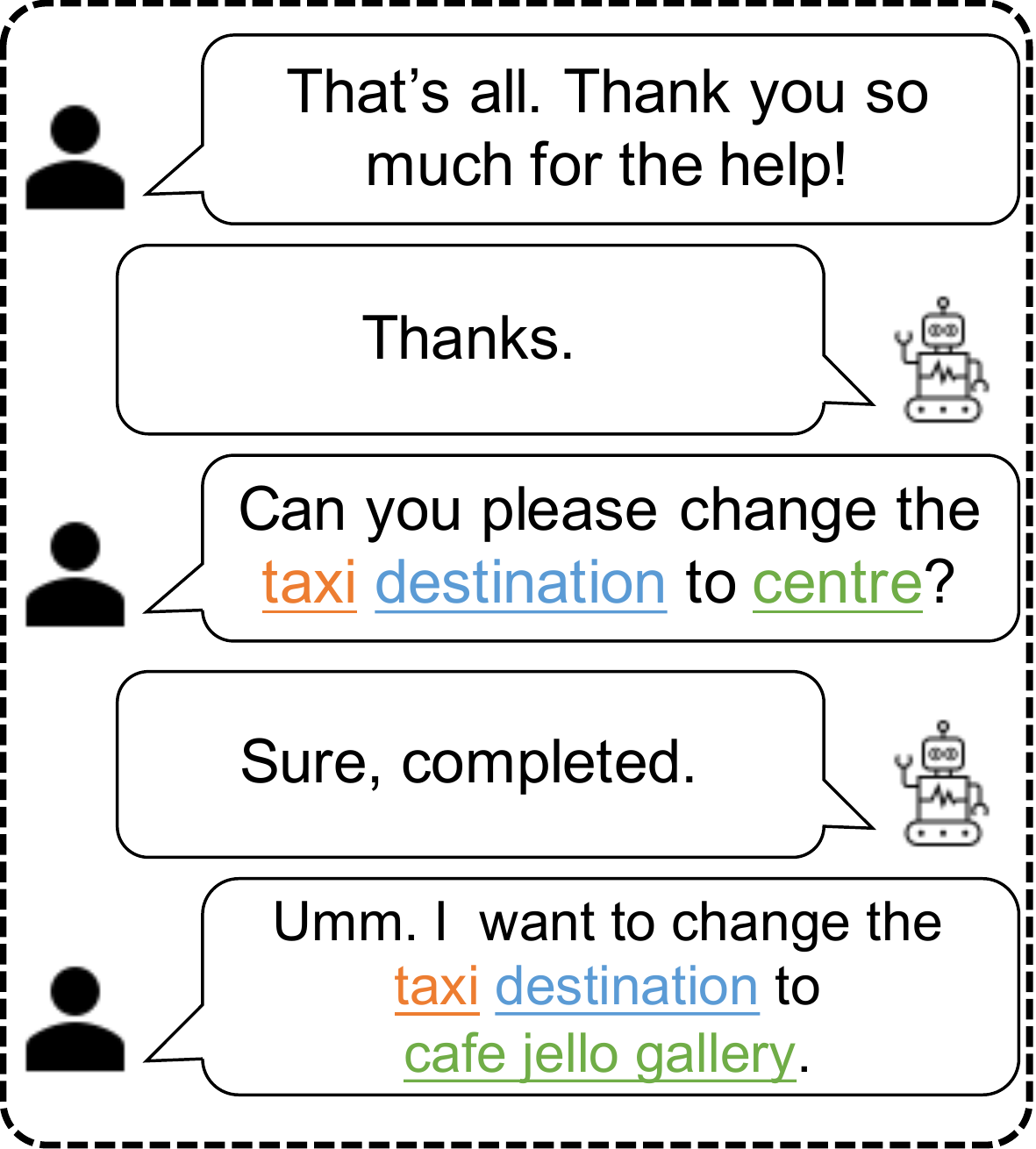}
        \caption{\dualval}
        \label{fig:3c}
    \end{subfigure}
    % \hspace{0.1cm}
    \begin{subfigure}[b]{0.245\textwidth}   
        \centering 
        \includegraphics[width=\textwidth]{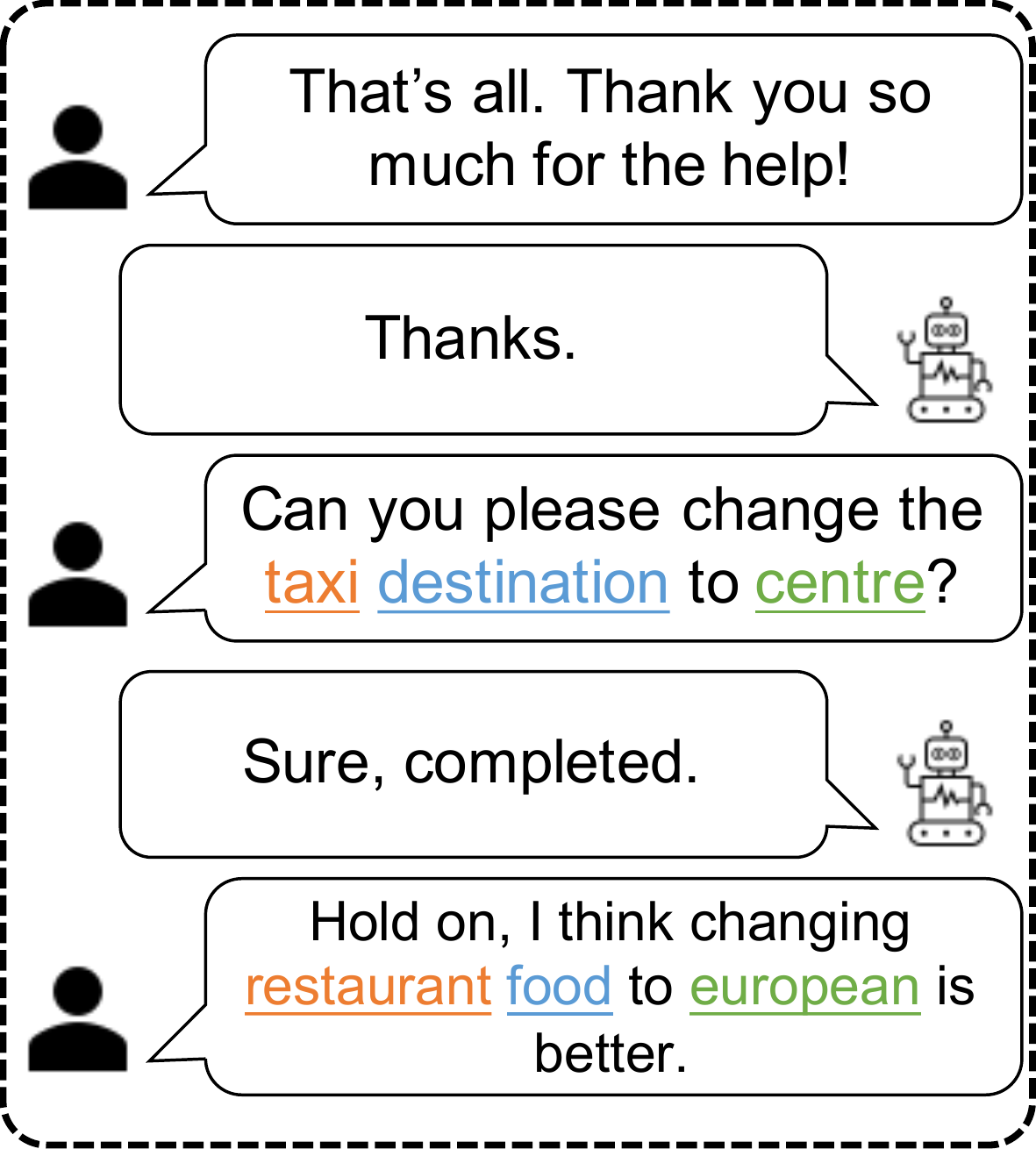}
        \caption{\dualslot}
        \label{fig:3d}
    \end{subfigure}
    \caption
    {An example of proposed turnback situations. Text in orange denotes a domain, blue denotes a slot, and green denotes a value.} 
    \label{fig:3}
\end{figure*}

\paragraph{Change my mind}
During a real conversation, people often change their minds. For example, when making a reservation for a restaurant, one might change the number of visitors, arrival time, or menu. When catching a taxi, the rider might ask the driver to go to their office first, and suddenly decide to go home to take a rest instead. Someone might want to sleep more, so they might delay their departure time. There are many other examples in which speakers change their mind or decision during a conversation. Unfortunately, the current well-known DST benchmark dataset does not seem to take these scenarios into serious consideration. All conversations continue naturally, and no one reverses what they have said. Some approaches reflect changing decisions of the user but only cover changes in the same dialog topic  \cite{bordes2017learning, mosig2020star}. Our contention regarding the conditions of a good DST benchmark dataset is that the conversations in the dataset should reflect more realistic situations, e.g., frequent turnback utterances, which are a main component of ordinary conversations in the real world.

This paper is partially related to \citet{conversational2022}, which points out the current task-oriented dialogue benchmark only considers short-term context rather than long history. Our turnback scenarios are the representative phenomena that show the lack of \textit{conversationality} of the benchmark dataset, defined in \citet{conversational2022}.

\begin{figure}
    \centering
    \includegraphics[width=\columnwidth]{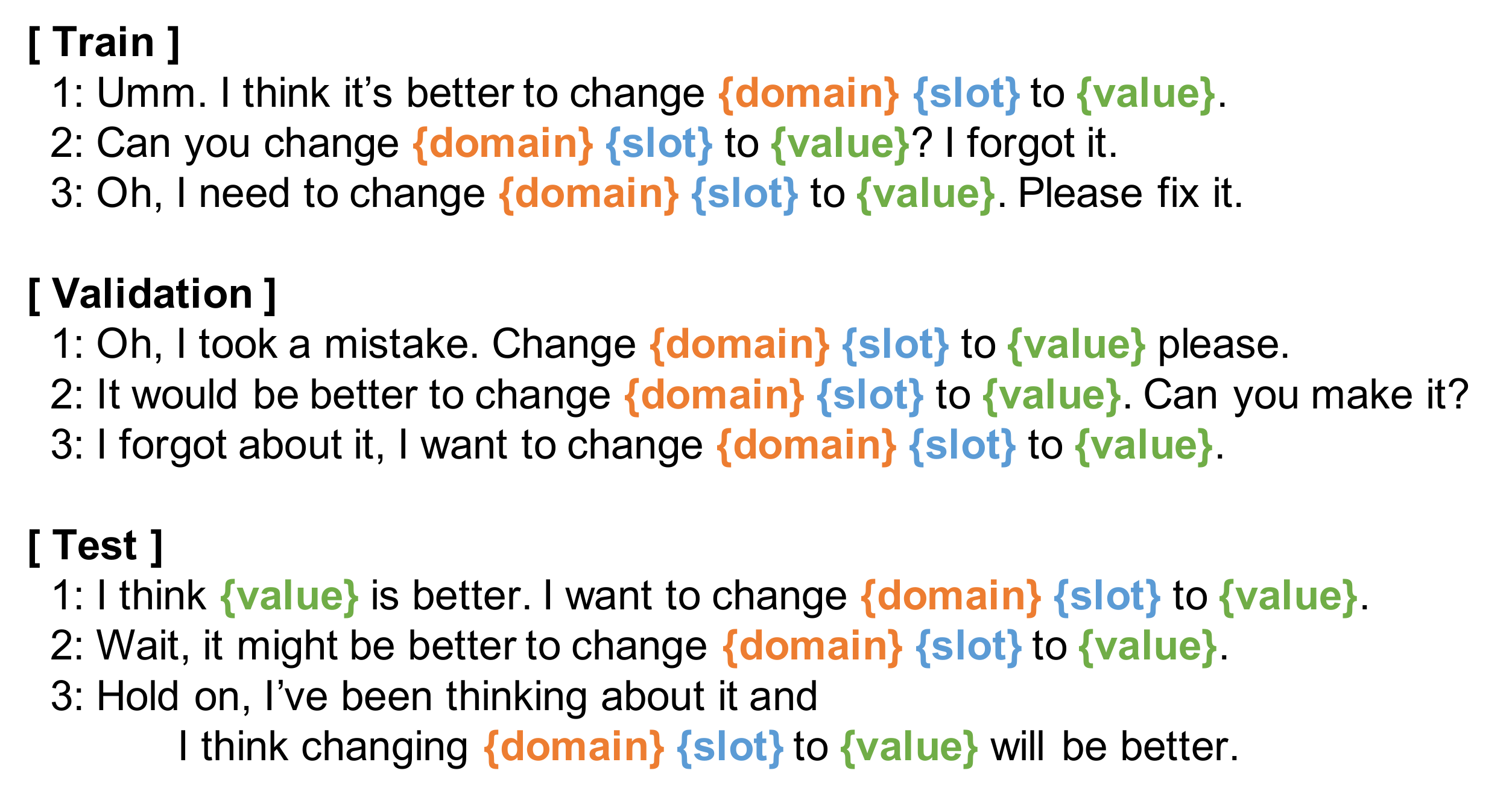}
    \caption
        {
        Template utterances of each phase (train, validation, and test).
        }
    \label{fig:2}
\end{figure}

\section{Method}
\label{method}

\begin{figure*}
    \centering
    \begin{subfigure}[b]{0.4\textwidth}
        \centering
        \includegraphics[width=\textwidth]{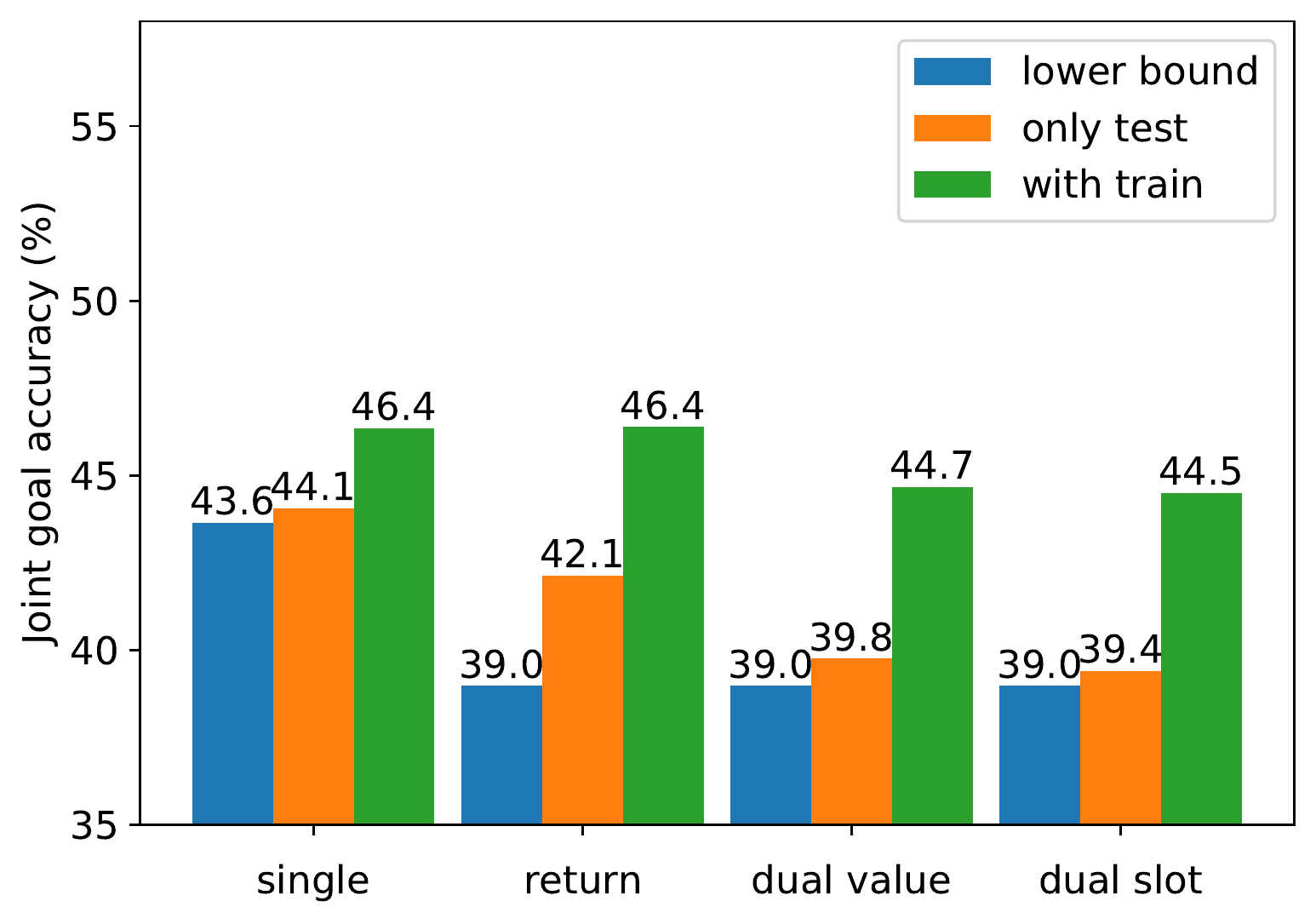}
        \caption{TRADE}
        \label{fig:5a}
    \end{subfigure}
    \hspace{0.7cm}
    \begin{subfigure}[b]{0.4\textwidth}  
        \centering 
        \includegraphics[width=\textwidth]{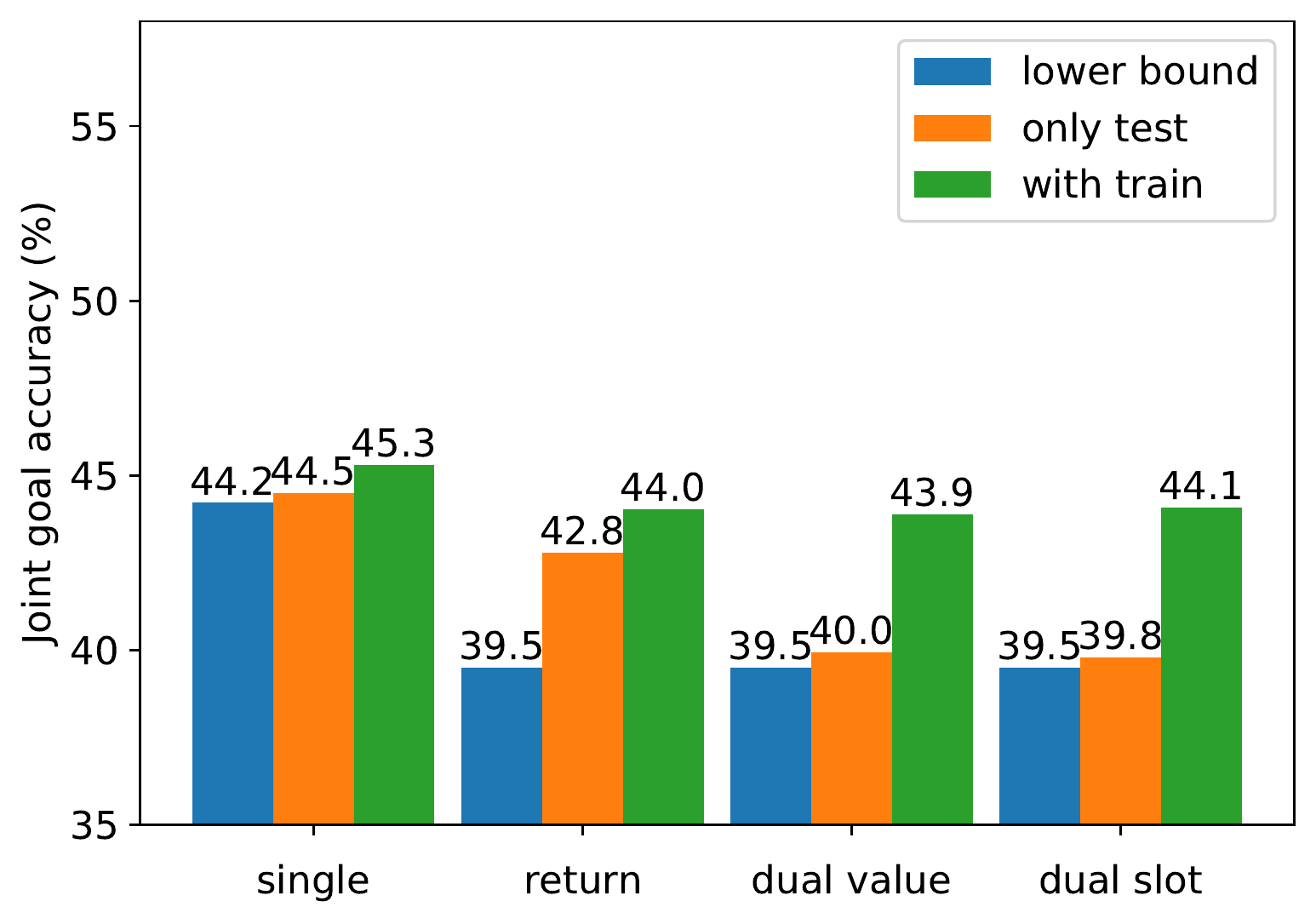}
        \caption{TRADE + CoCo}
        \label{fig:5b}
    \end{subfigure}
    % \vskip\baselineskip
    \begin{subfigure}[b]{0.4\textwidth}   
        \centering 
        \includegraphics[width=\textwidth]{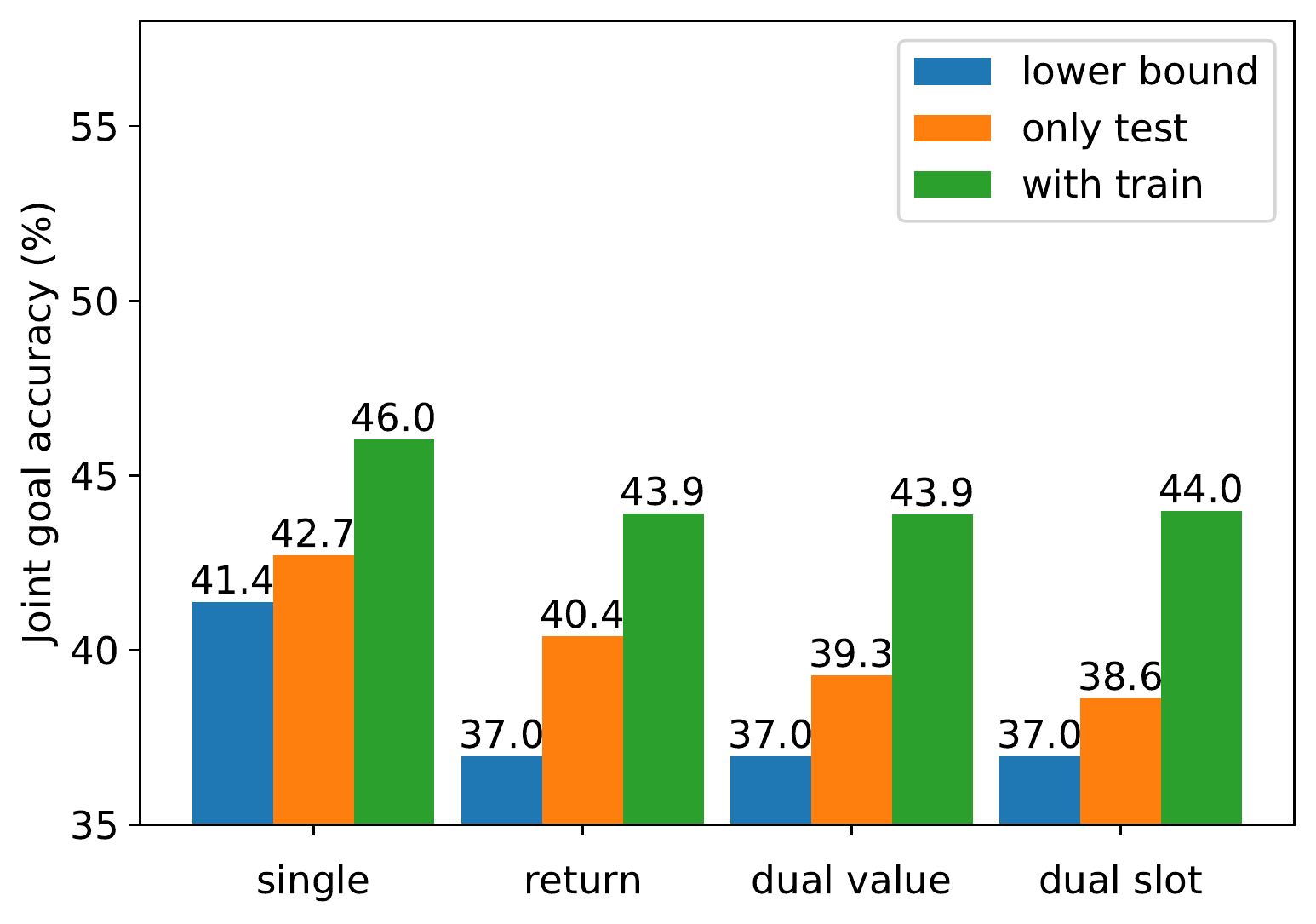}
        \caption{SUMBT}
        \label{fig:5c}
    \end{subfigure}
    \hspace{0.7cm}
    \begin{subfigure}[b]{0.4\textwidth}   
        \centering 
        \includegraphics[width=\textwidth]{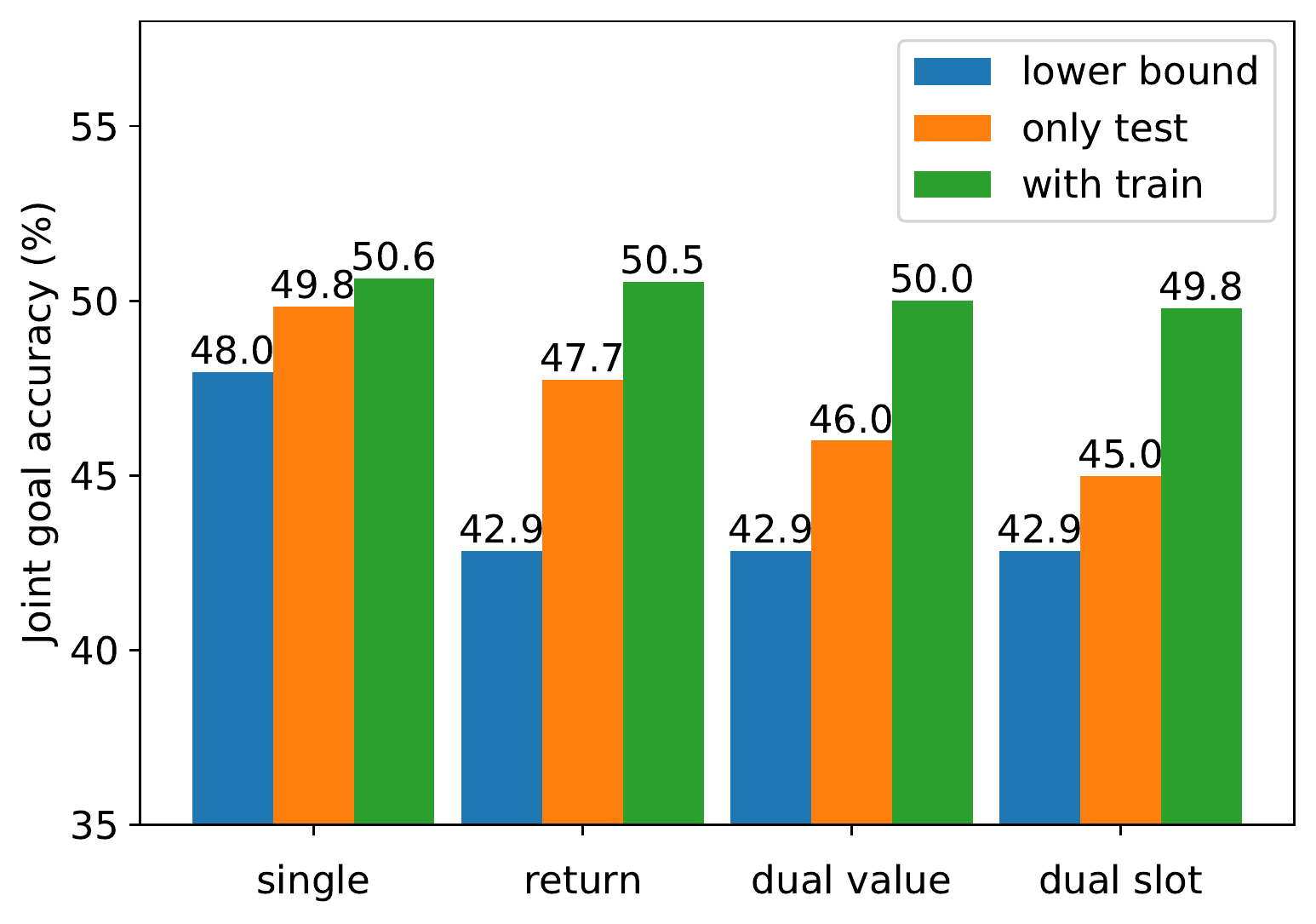}
        \caption{Transformer-DST}
        \label{fig:5d}
    \end{subfigure}
    \caption
    {Performance gap based on the existence of turnback in the training data. Lower bound indicates the performance of not correctly predicting additional turnback turns at all.} 
    \label{fig:5}
\end{figure*}

To test whether the model trained with the current DST dataset can track the change in value of the turnback situation, we assume four turnback scenarios and inject these turnback utterances at the end of every dialogue, as represented as Figure \ref{fig:3}. In other words, each data containing dialogue of $t$ turns can be formulated as $ X_t=\{ (U_1^{sys}, U_1^{usr}), \dots, (U_t^{sys}, U_t^{usr}) \} $, and we then append an extra template-generated turn with one of the aforementioned turnback situations at the end of the existing data, resulting in $ X_k=\{ (U_1^{sys}, U_1^{usr}), \dots, (U_k^{sys}, U_k^{usr}) \} $, where $k=t+1$ for a single turnback situation or $k=t+2$ for multiple situations. Figure \ref{fig:2} shows examples of a turnback used in each dataset. Note that we used different templates for different datasets to avoid an overlap across the datasets. Whenever applying a template-based utterance generation, the arbitrary template of each phase is selected at each turn of dialogue. 

As the main purpose of this paper is to investigate whether the model can follow the user's mind-changing utterances, we designed the simplest form of turnback utterances: injecting them to the last turn and generating utterances using templates. The former is to assume the mind-changing within the longest history in a single dialogue, and the latter is to show that models cannot track changing values when even the most informative turnback utterances are explicitly provided. Accordingly, we defined four variants of turnback situations as follows:

\paragraph{\single}
Users change the value of a particular slot only once, as shown in Figure \ref{fig:3a}. Basically, a single turnback utterance is constructed using the last turn of the dialogue because it contains accumulated belief states that appeared throughout the dialogue. Figure \ref{fig:4} shows the process of generating a single turnback utterance and skipping the process when there is no belief stated during the dialogue.

\paragraph{\return}
Users change the value of a particular slot but return to the original value again, as shown in Figure \ref{fig:3b}. This means that the final belief state after injecting a return turnback utterance is the same as the belief state of the original dataset. In this case, the first turnback utterance can be generated like a single turnback process, and the second turnback utterance is then generated identically by simply replacing the changed value with the original value.

\paragraph{\dualval}
Users sequentially change the value of a particular slot twice, as shown in Figure \ref{fig:3c}. Dual value turnback utterances can be generated in the same way as return turnback utterances, but can be generalized to a triple or quadruple value turnback if there are more than two available values in the slot on the ontology.

\paragraph{\dualslot}
Users first change the value of a particular slot and then also change the value of a different slot, as represented in Figure \ref{fig:3d}. This can be generated simply by applying a single turnback twice; however, there must be more than two total belief states to apply this scenario.

\section{Experiments}
\label{experiments}

\subsection{Experimental setup}
\label{exp_setup}

We verified our hypothesis using the MultiWOZ 2.1 \cite{eric2019multiwoz}, the most commonly used DST dataset in previous studies. As a performance metric, the joint goal accuracy was employed. The joint goal accuracy is a standard criterion used to check if the model tracks the triplet of \texttt{(domain, slot, value)} precisely. When tracked correctly, the joint goal accuracy is marked as 1, and is otherwise 0. The numbers of training, validation, and test sets are 8420, 1000, and 999, respectively. The open-source code for the TRADE model was from CoCo repository\footnote{https://github.com/salesforce/coco-dst}, while the code for SUMBT\footnote{https://github.com/SKTBrain/SUMBT} and Transformer-DST\footnote{https://github.com/zengyan-97/Transformer-DST} was from the original author, respectively. For the TRADE model, we also considered the model trained jointly with CoCo-augmented dataset \cite{li2020coco}.  All the experiments explained later were conducted using a machine with the NVIDIA GeForce RTX 3090 GPU.

\subsection{Main results}

Figure \ref{fig:5} shows the main results. As the extra turnback utterances are appended to the original dataset, we reported the performance lower bound where the model does not predict the additional states of turnback utterances at all (blue color). In other words, the joint goal accuracy of every turn of turnback scenarios is zero in the lower bound setting. The performance of the original model with turnback-included test set is reported with the orange color. Compared to the lower bound, the model trained with the original set correctly predicts only a few altered dialogue states. In the case of multiple turnbacks (i.e., \return, \dualval, and \dualslot), the models with
\return resulted in relatively better performance than the others. This is not because the model predict the state values in the turnback utterance correctly, but because \return has the same value with the original state value. Note that these turnback utterances generated using templates are the \textit{easiest} form of the situation, explicitly providing the entire information of \texttt{domain}, \texttt{slot}, and \texttt{value}.

\begin{table*}
    \centering
    \begin{tabularx}{\textwidth}{c>{\raggedright}X}
        \toprule
        \textbf{Turn\ \#} & \textbf{Dialogue History} \tabularnewline
        \midrule
        1 & System: `` '' \\
            User: ``I need a taxi. I'll be departing from la raza.'' \tabularnewline
        \midrule
        2 & System: ``I can help you with that. When do you need to leave?'' \\
            User: ``I would like to leave after 11:45 please.'' \tabularnewline
        \midrule
        3 & System: ``Where will you be going?'' \\
            User: ``I'll be going to restaurant 17.'' \tabularnewline
        \midrule
        4 & System: ``I have booked for you a black volkswagen, the contact number is 07552762364. Is there anything else I can help you with?'' \\
            User: ``No, that's it. Thank you!'' \tabularnewline
        \midrule
        5 & System: ``Completed.'' \\
            User: ``Wait , it might be better to change \textbf{taxi} \textbf{leave at} to \textbf{15:00}.'' \tabularnewline
        \midrule
        6 & System: ``Sure. Anything else?'' \\
            User: ``Hold on , I've been thinking about it and I think changing \textbf{taxi} \textbf{destination} to \textbf{finches bed and breakfast} will be better.'' \tabularnewline
        \bottomrule
    \end{tabularx}
    \caption{Sample dialogue of test set with additional \dualslot situation (SNG01367.json).}
    \label{table:2}
\end{table*}

\begin{table*} 
    \begin{center}
    \setlength{\tabcolsep}{10pt}
        \begin{tabularx}{\textwidth}{ccc}
            \toprule
            \textbf{Gold state} & \textbf{Predicted state} & \textbf{Predicted state} \\
            \textbf{(label)} & \textbf{(original model)} & \textbf{(\textsc{dual-slot}-trained model)} 
            \tabularnewline
            \midrule
            "taxi-departure-la raza", & "taxi-departure-la raza",  & "taxi-departure-la raza", \tabularnewline 
             "taxi-leaveat-11:45", & "taxi-leaveat-11:45", & "taxi-leaveat-11:45", \tabularnewline 
             "taxi-destination- & "taxi-destination- & "taxi-destination- \tabularnewline
             restaurant 17" & restaurant 17" & restaurant 17"
             \tabularnewline 
            \midrule
            "taxi-departure-la raza", & "taxi-departure-la raza",  & "taxi-departure-la raza", \tabularnewline 
             "taxi-leaveat-\textbf{15:00}", & "taxi-leaveat-\underline{11:45}", & "taxi-leaveat-\textbf{15:00}",  \tabularnewline 
            "taxi-destination- & "taxi-destination- & "taxi-destination- \tabularnewline
             restaurant 17" & restaurant 17" & restaurant 17" \tabularnewline
            \midrule
            "taxi-departure-la raza", & "taxi-departure-la raza",  & "taxi-departure-la raza", \tabularnewline 
             "taxi-leaveat-\textbf{15:00}", & "taxi-leaveat-\underline{11:45}", & "taxi-leaveat-\textbf{15:00}",  \tabularnewline 
            "taxi-destination- & "taxi-destination- & "taxi-destination- \tabularnewline 
             \textbf{finches bed and breakfast}" & \textbf{finches bed and breakfast}" & \textbf{finches bed and breakfast}"  \tabularnewline 
            \bottomrule
        \end{tabularx}
    \caption{The model prediction on \dualslot situation at turn 4, 5, and 6 (SNG01367.json).}
    \label{table:3}
    \end{center}
\end{table*}

\begin{table*}
    \centering
    \begin{tabular}{lcccccc}
        \toprule
        \multicolumn{5}{c}{\hfill \single}                   \\
        \cmidrule(r){2-6}
         & Train-0\%     & Train-30\% & Train-50\% & Train-70\% & Train-100\% & \multicolumn{1}{c}{Difference}  \\
        \midrule
        Test-0\% & \textbf{54.47}  & 54.40  & 54.32  & \underline{54.44}& 52.80 & -0.03\%p \\
        Test-30\%    & 53.04  & 53.81 &\underline{53.84} & \textbf{54.00}  &  52.22 & 0.96\%p  \\
        Test-50\%   & 52.06  & \underline{53.44} &  53.36&  \textbf{53.46}&  51.88 & 1.40\%p \\
        Test-70\%     & 50.90  & \textbf{52.81} & \underline{52.78} & 52.73 &  51.12 & 1.91\%p \\
        Test-100\%     & 49.84  & 51.98 &\underline{52.23}  & \textbf{52.32} & 50.65 & 2.48\%p  \\ 
        \bottomrule
        \multicolumn{5}{l}{\footnotesize  * Bold denotes the best, and underline denotes the second-best performance.} \\
    \end{tabular}
    \caption{Joint goal accuracy (\%) of Transformer-DST with different \single proportions.}
    \label{table:4}
\end{table*}

\subsection{Including turnback dialogues in the training set}

Because the main hypothesis was sufficiently supported by the first experiment, we further investigated whether including turnback situations in the training dataset can prevent the model from not being able to trace the changing values. We inserted turnback utterances at the end of all training train and validation data, and different template utterances were randomly used for the training and validation phases, as illustrated in Figure \ref{fig:2}. 

The green-colored bar in Figure \ref{fig:5} shows the joint goal accuracy for each turnback scenario before and after the turnback utterance are included in the training and validation datasets with a performance lower bound of newly added turnback turns. The performance always improves irrespective of the turnback scenarios and DST models. Also note that the performance recovery is more significant for more complicated turnback scenarios. Injecting turnback utterances increases the joint goal accuracy by 1.83\%p on average for a single turnback, whereas the average improvement is 4.90\%p for the dual slot turnback.

In addition to achieving a quantitative rebound in performance, we also conducted a qualitative comparison of the model predictions before and after the turnback injection in the training and validation datasets. Table \ref{table:2} shows an example of dual slot turnback dialogue, and the predicted states of the Transformer-DST model are as shown in Table \ref{table:3}. The prediction results of the remaining three turnback situations are also provided in Tables \ref{table:a1}, \ref{table:a2}, and \ref{table:a3}.
The first row of Table \ref{table:3} is the last turn of the original dialogue, and we can see that both the original and dual-trained model predict the belief states correctly. In the second and third rows of the same table, when the values of two slots are sequentially changed, the original model can catch only one changing value (\textit{`finches bed and breakfast'}). Not being able to follow all changes is frequently detected with the original model in other test dialogues. By contrast, the model trained with the turnback utterances can correctly predict the entire belief state, as shown in the last row and the last column of Table \ref{table:3}. 

Based on the results shown in Figure \ref{fig:5} and Table \ref{table:3}, we can conclude that the performance degeneration of the DST models is not because the DST model structures are incorrect but because they do not have a chance to train such turnback utterances with the current benchmark DST dataset, which means that the MultiWOZ dataset does not have a sufficient coverage yet for dialogues in the real-world.

\subsection{Difference in performance according to turnback proportion}

We also conducted an ablation study on how the turnback utterance proportions in the training and test dataset affect the DST performance. We evaluate five different proportions of turnback-injected training and test datasets (i.e., 0\%, 30\%, 50\%, 70\%, and 100\%) with corresponding turnback-test situations, resulting in a total of 25 combinations of training-test turnback proportions. We named each turnback-mixed dataset \textit{phase-N\%}. For example, Train-30\% denotes the dataset in which 30\% of the turnback utterances are applied to the existing dialogues, and the remaining 70\% of the original dialogues are unmodified. The performances of Transformer-DST are shown in Table \ref{table:4}. The performance of the other models are provided in Tables \ref{table:a4}, \ref{table:a5}, and \ref{table:a6}. The last column of the table is the difference between the best-proportion model performance and the original performance.

Based on Table \ref{table:4}, we can draw the following observations. First, adding moderate turnback utterances does not significantly affect the performance on Test-0\%, which is the original test dataset. The joint goal accuracies of Train-30\%, Train-50\%, and Train-70\% are very close to that of Train-0\%. Second, high proportions of turnback utterances in the training set help recover the performance in most cases. With regard to turnback ratio in the training dataset, above 70\% of the turnback utterance show the best performance in Table \ref{table:4}, \ref{table:a4}, and \ref{table:a6}. In the case of Table \ref{table:a5}, we expect that conterfactual slot combinations provided in CoCo-augmented dataset can assist the model's robust prediction.

\section{Conclusion}
\label{conclusion}

A DST model should focus on properly reacting to unpredictable scenarios from a human speaker.
From this perspective, using realistic benchmark datasets for the model is crucial. To validate recent DST models trained on the commonly used DST benchmark dataset, we first designed a template-based (but enough to verify the hypothesis) data injection method to create a turnback situation and modified the test dataset by appending one of four trunback scenarios to the end of the dialogue. Our experiment showed that the current model trained using the existing benchmark cannot track the changing values well when users change their decisions. We also conducted additional experiment to investigate whether the model performance can be recovered if the turnback utterances are properly included in the training dataset. Experimental results showed that the joint goal accuracy was improved for all turnback scenarios when the models were trained on the dataset with turnback utterances. The ablation study shows that moderately including the turnback utterances can manage a broader range of turnback proportions. Our experimental results emphasize that constructing a right benchmark dataset is as important as developing an advanced model structure in NLP tasks.

Despite the meaningful results, we argue that the turnback utterance is just one of many situations that can happen in a real-world conversation. If more diverse realistic dialogue scenarios are reflected in the DST benchmark dataset, the bias of models trained on it can be significantly reduced.

{
\small

\section*{Acknowledgment}

This research was supported and funded by the Korean National Police Agency. [Pol-Bot Development for Conversational Police Knowledge Services / PR09-01-000-20]

\bibliographystyle{acl_natbib}
\bibliography{references}

}

\appendix

\section{Appendix}

\begin{figure*}
    \centering
    \setcounter{figure}{0}
    \renewcommand{\thefigure}{A\arabic{figure}}
    \includegraphics[width=\textwidth]{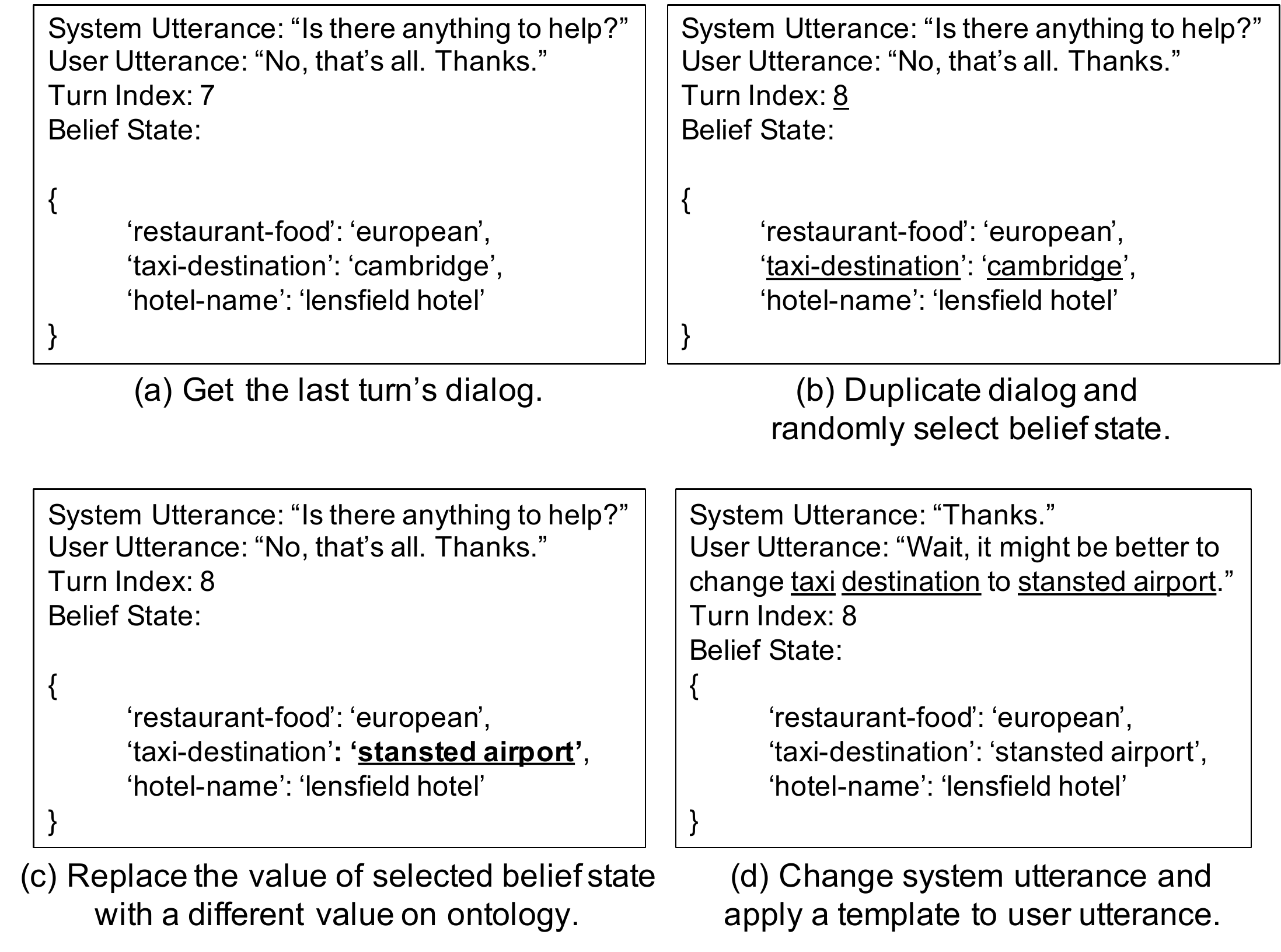}
    \caption
        {
        Process of \single dialogue generation.
        }
    \label{fig:4}
\end{figure*}

\begin{table*}[h!]
    \begin{center}
    \setcounter{table}{0}
    \renewcommand{\arraystretch}{0.9}
    \renewcommand{\thetable}{A\arabic{table}}
    \setlength{\tabcolsep}{18pt}
        \begin{tabularx}{\textwidth}{ccc}
            \toprule
            \textbf{Gold state} & \textbf{Predicted state} & \textbf{Predicted state} \\
            \textbf{(label)} & \textbf{(original model)} & \textbf{(\textsc{single}-trained model)}
            \tabularnewline
            \midrule
            "taxi-departure-la raza", & "taxi-departure-la raza",  & "taxi-departure-la raza", \tabularnewline 
             "taxi-leaveat-11:45", & "taxi-leaveat-11:45", & "taxi-leaveat-11:45", \tabularnewline 
             "taxi-destination- & "taxi-destination & "taxi-destination \tabularnewline
             restaurant 17" & restaurant 17" & restaurant 17"
             \tabularnewline 
            \midrule
            "taxi-departure-, & "taxi-departure-  & "taxi-departure- \tabularnewline 
            \textbf{london liverpool street}", & \underline{la raza}", & \textbf{london liverpool street}",  \tabularnewline 
             "taxi-leaveat-11:45", & "taxi-leaveat-11:45", & "taxi-leaveat-11:45", \tabularnewline 
            "taxi-destination- & "taxi-destination- & "taxi-destination- \tabularnewline
             restaurant 17" & restaurant 17" & restaurant 17" \tabularnewline
            \bottomrule
        \end{tabularx}
    \caption{Model prediction on \single situation at turns 4 and 5 (SNG01367.json).}
    \label{table:a1}
    \end{center}
\end{table*}

\begin{table*}[h!]
    \begin{center}
    \renewcommand{\arraystretch}{0.9}
    \renewcommand{\thetable}{A\arabic{table}}
    \setlength{\tabcolsep}{18pt}
        \begin{tabularx}{\textwidth}{ccc}
            \toprule
            \textbf{Gold state} & \textbf{Predicted state} & \textbf{Predicted state} \\
            \textbf{(label)} & \textbf{(original model)} & \textbf{(\textsc{return}-trained model)}
            \tabularnewline
            \midrule
            "taxi-departure-la raza", & "taxi-departure-la raza",  & "taxi-departure-la raza", \tabularnewline 
             "taxi-leaveat-11:45", & "taxi-leaveat-11:45", & "taxi-leaveat-11:45", \tabularnewline 
             "taxi-destination- & "taxi-destination- & "taxi-destination- \tabularnewline
             restaurant 17" & restaurant 17" & restaurant 17"
             \tabularnewline 
            \midrule
            "taxi-departure-, & "taxi-departure-  & "taxi-departure- \tabularnewline 
            \textbf{the copper kettle}", & \underline{la raza}", & \textbf{the copper kettle}",  \tabularnewline 
             "taxi-leaveat-11:45", & "taxi-leaveat-11:45", & "taxi-leaveat-11:45", \tabularnewline 
            "taxi-destination- & "taxi-destination- & "taxi-destination- \tabularnewline
             restaurant 17" & restaurant 17" & restaurant 17" \tabularnewline
            \midrule
            "taxi-departure-\textbf{la raza}", & "taxi-departure-\textbf{la raza}",  & "taxi-departure-\textbf{la raza}", \tabularnewline 
             "taxi-leaveat-11:45", & "taxi-leaveat-11:45", & "taxi-leaveat-11:45", \tabularnewline 
            "taxi-destination- & "taxi-destination- & "taxi-destination- \tabularnewline 
             restaurant 17" & restaurant 17" & restaurant 17" \tabularnewline
            \bottomrule
        \end{tabularx}
    \caption{Model prediction on \return situation at turns 4, 5, and 6 (SNG01367.json).}
    \label{table:a2}
    \end{center}
\end{table*}

\begin{table*}[h!]
    \begin{center}
    \renewcommand{\arraystretch}{0.9}
    \renewcommand{\thetable}{A\arabic{table}}
    \setlength{\tabcolsep}{18pt}
        \begin{tabularx}{\textwidth}{ccc}
            \toprule
            \textbf{Gold state} & \textbf{Predicted state} & \textbf{Predicted state} \\
            \textbf{(label)} & \textbf{(original model)} & \textbf{(\textsc{dual-value}-trained model)}
            \tabularnewline
            \midrule
            "taxi-departure-la raza", & "taxi-departure-la raza",  & "taxi-departure-la raza", \tabularnewline 
             "taxi-leaveat-11:45", & "taxi-leaveat-11:45", & "taxi-leaveat-11:45", \tabularnewline 
             "taxi-destination- & "taxi-destination- & "taxi-destination- \tabularnewline
             restaurant 17" & restaurant 17" & restaurant 17"
             \tabularnewline 
            \midrule
            "taxi-departure-la raza", & "taxi-departure-la raza",  & "taxi-departure-la raza", \tabularnewline 
             "taxi-leaveat-\textbf{10:15}", & "taxi-leaveat-\textbf{10:15}", & "taxi-leaveat-\textbf{10:15}",  \tabularnewline 
            "taxi-destination- & "taxi-destination- & "taxi-destination- \tabularnewline
             restaurant 17" & restaurant 17" & restaurant 17" \tabularnewline
            \midrule
            "taxi-departure-la raza", & "taxi-departure-la raza",  & "taxi-departure-la raza", \tabularnewline 
             "taxi-leaveat-\textbf{12:00}", & "taxi-leaveat-\underline{10:15}", & "taxi-leaveat-\textbf{12:00}", \tabularnewline 
            "taxi-destination- & "taxi-destination- & "taxi-destination- \tabularnewline 
             restaurant 17" & restaurant 17" & restaurant 17" \tabularnewline
            \bottomrule
        \end{tabularx}
    \caption{Model prediction on \dualval situation at turn 4, 5, and 6 (SNG01367.json).}
    \label{table:a3}
    \end{center}
\end{table*}

\begin{table*}[h!]
    \renewcommand{\thetable}{A\arabic{table}}
    \centering
    \begin{tabular}{lcccccc}
        \toprule
        \multicolumn{5}{c}{\hfill \single}                   \\
        \cmidrule(r){2-6}
         & Train-0\%     & Train-30\% & Train-50\% & Train-70\% & Train-100\% & \multicolumn{1}{c}{Difference}  \\
        \midrule
        Test-0\% & \textbf{49.55}  & 48.47  & 48.25  & 48.11 & \underline{48.81} & -0.74\%p \\
        Test-30\%    & \textbf{47.82}  & 47.41 & 47.16 & 47.16  &  \textbf{47.82} & 0.00 \%p  \\
        Test-50\%   & 46.52  & 46.62 & 46.41  & \underline{46.67} &  \textbf{47.24} & 0.72\%p \\
        Test-70\%     & 45.31  & \underline{45.92} & 45.63 & 45.85 &  \textbf{46.50} & 1.19\%p \\
        Test-100\%     & 44.05  & 45.12 & 45.13 & \underline{45.29} & \textbf{46.36} & 2.31\%p \\
        \bottomrule
        \multicolumn{5}{l}{\footnotesize  * Bold denotes the best, and underline denotes the second-best performance.} \\
    \end{tabular}
    \caption{Joint goal accuracy (\%) of TRADE with different \single proportions.}
    \label{table:a4}
\end{table*}

\begin{table*}[h!]
    \renewcommand{\thetable}{A\arabic{table}}
    \centering
    \begin{tabular}{lcccccc}
        \toprule
        \multicolumn{5}{c}{\hfill \single}                   \\
        \cmidrule(r){2-6}
         & Train-0\%     & Train-30\% & Train-50\% & Train-70\% & Train-100\% & \multicolumn{1}{c}{Difference}  \\
        \midrule
        Test-0\% & \textbf{50.21}  & 48.40  & \underline{49.80}  & 47.73 & 48.05 & -0.41\%p \\
        Test-30\%    & \underline{48.36}  & 47.30 & \textbf{48.74} & 46.81  &  47.22 & 0.38\%p  \\
        Test-50\%   & \underline{47.13}  & 46.57 & \textbf{48.16}  & 46.07 &  46.62 & 1.03\%p \\
        Test-70\%     & \underline{46.02}  & 45.57 & \textbf{47.42} & 45.38 &  45.89 & 1.40\%p \\
        Test-100\%     & 44.49  & 44.75 & \textbf{46.73} & 44.75 & \underline{45.30} & 2.24\%p \\
        \bottomrule
        \multicolumn{5}{l}{\footnotesize  * Bold denotes the best, and underline denotes the second-best performance.} \\
    \end{tabular}
    \caption{Joint goal accuracy (\%) of TRADE + CoCo with different \single proportions.}
    \label{table:a5}
\end{table*}

\begin{table*}[h!]
    \renewcommand{\thetable}{A\arabic{table}}
    \centering
    \begin{tabular}{lcccccc}
        \toprule
        \multicolumn{5}{c}{\hfill \single}                   \\
        \cmidrule(r){2-6}
         & Train-0\%     & Train-30\% & Train-50\% & Train-70\% & Train-100\% & \multicolumn{1}{c}{Difference}  \\
        \midrule
        Test-0\% & 46.99  & 46.24  & 46.32  & \textbf{47.16} & \underline{47.10} & 0.17\%p  \\
        Test-30\%    & 45.59  & 46.57 & 46.17 & \underline{47.18}  &  \textbf{47.38} & 1.79\%p \\
        Test-50\%   & 44.80  & 46.29 & 45.70  & \underline{46.70} &  \textbf{47.22} & 2.42\%p \\
        Test-70\%     & 43.73  & 45.54 & 45.13 & \underline{46.11} &  \textbf{46.39} & 2.66\%p \\
        Test-100\%     & 42.72  & 45.01 & 44.70 & \underline{45.62} & \textbf{46.04 } & 3.32\%p \\
        \bottomrule
        \multicolumn{5}{l}{\footnotesize  * Bold denotes the best, and underline denotes the second-best performance.} \\
    \end{tabular}
    \caption{Joint goal accuracy (\%) of SUMBT with different \single proportions.}
    \label{table:a6}
\end{table*}

\end{document}